\documentclass[journal]{IEEEtran}

\usepackage{xcolor,soul,framed} 

\colorlet{shadecolor}{yellow}
\usepackage[pdftex]{graphicx}
\usepackage[cmex10]{amsmath}
\usepackage{array}
\usepackage{mdwmath}
\usepackage{mdwtab}
\usepackage{eqparbox}
\usepackage{url}
\usepackage{comment}
\usepackage{romannum}
\usepackage{xcolor}
\usepackage{hyperref}
\hypersetup{hidelinks=true}
\usepackage{algorithm,algorithmic}
\usepackage[caption=false]{subfig}

\usepackage{amssymb}

\begin{document}
\title{ViT LoS V2X: Vision Transformers for Environment-aware LoS Blockage Prediction for 6G Vehicular Networks}

\author{Ghazi Gharsallah, \IEEEmembership{Student Member, IEEE} and Georges Kaddoum, \IEEEmembership{Senior Member, IEEE}

Electrical Engineering Department, École de Technologie Supérieure, Montreal, Canada.}

\maketitle

\begin{abstract}
As wireless communication technology progresses towards the sixth generation (6G), high-frequency millimeter-wave (mmWave) communication has emerged as a promising candidate for enabling vehicular networks. It offers high data rates and low-latency communication. However, obstacles such as buildings, trees, and other vehicles can cause signal attenuation and blockage, leading to communication failures that can result in fatal accidents or traffic congestion. Predicting blockages is crucial for ensuring reliable and efficient communications.
Furthermore, the advent of 6G technology is anticipated to integrate advanced sensing capabilities, utilizing a variety of sensor types. These sensors, ranging from traditional RF sensors to cameras and Lidar sensors, are expected to provide access to rich multimodal data, thereby enriching communication systems with a wealth of additional contextual information. Leveraging this multimodal data becomes essential for making precise network management decisions, including the crucial task of blockage detection.
In this paper, we propose a Deep Learning (DL)-based approach that combines Convolutional Neural Networks (CNNs) and customized Vision Transformers (ViTs) to effectively extract essential information from multimodal data and predict blockages in vehicular networks. Our method capitalizes on the synergistic strengths of CNNs and ViTs to extract features from time-series multimodal data, which include images and beam vectors. To capture temporal dependencies between the extracted features and the blockage state at future time steps, we employ a Gated Recurrent Unit (GRU)-based architecture.
We train and evaluate our proposed method on the DL dataset framework for vision-aided wireless communications (ViWi) and demonstrate its potential for predicting blockages in vehicular networks through simulations. Our results show that the proposed approach achieves high accuracy and outperforms state-of-the-art solutions, achieving more than $95\%$ accurate predictions. This contribution is expected to advance the development of ultra-reliable and low-latency communication in vehicular networks.
\end{abstract}


\IEEEpeerreviewmaketitle

\section{Introduction}
\subsection{Background and Motivation}
\IEEEPARstart{W}{ireless} communication has shown one of the fastest growth rates in the previous decades, with the commercial deployment of the fifth-generation wireless communication (5G) in 2020. 5G is a revolutionary technology that offers ultra-reliable and low-latency communications (uRLLC), as well as enhanced mobile broadband (eMBB) services. 
Researchers are now turning their attention towards sixth-generation (6G) wireless communication systems, which aim to offer "connected intelligence" instead of "connected things" in various sectors \cite{8808168, 8869705, 9349624, 9426946}.
The transition to 6G demands higher data rates and channel capacity, requiring the use of higher frequencies, including millimeter-wave (mmWave) and sub-terahertz communications \cite{8732419} \cite{9390169}.

However, these frequencies present new challenges, such as their vulnerability to blockage which results in signal-to-noise ratio (SNR) dips and frequent disconnections. Therefore, Line-of-Sight (LoS) and Non-Line-of-Sight (NLoS) predictions are vital for future high-frequency communication applications, where we can predict if the future connection between a base station and a given user will be blocked or not, especially in highly dynamic networks, such as vehicular networks, which are one of the most vulnerable applications to user disconnections.

With the development of 6G sensing and the availability of various sensor types, such as high-resolution cameras, LiDAR, radar, and acoustic sensors \cite{9376324, 9737357}. These sensors facilitate the collection of diverse data types, including visual imagery, depth information, and acoustic signals, thereby providing a comprehensive understanding of the surrounding environment. This data represents a significant opportunity to enhance network management decisions, including the crucial task of blockage detection. Leveraging this multimodal data is essential for ensuring the reliability and efficiency of vehicular communication networks.

The vehicle-to-everything (V2X) technology, which enables vehicles to communicate with other vehicles, infrastructure, and road users, is a key enabler of intelligent transportation systems \cite{10421061} \cite{9831302}.
The primary aim of V2X communication is to improve road safety, traffic efficiency, and driver experience by providing real-time information about road conditions, traffic flow, and nearby vehicles. 
However, V2X networks face significant challenges, such as high mobility, intermittent connectivity, and limited resources, which affect the performance of the network.
The complexity introduced by these challenges will create highly intricate vehicular network scenarios, making it extremely difficult, if not impossible, to model such systems for LoS prediction using traditional probabilistic methods and optimization models \cite{7143330, 8166758, 9121328, 8968748, 02723}.
Contextualization in V2X communication involves the integration and analysis of environmental, vehicular, and network contexts to enhance communication reliability and efficiency \cite{10242012}. This approach is particularly relevant to LoS prediction solutions in V2X networks by understanding the context in which vehicles operate, including dynamic changes in the environment and vehicle behavior.

The introduction of advanced machine learning architectures, such as Convolutional Neural Networks (CNNs) and Vision Transformers (ViTs) \cite{11929}, into the realm of vehicular communications opens new avenues for addressing these challenges. CNNs, known for their ability to capture spatial patterns and correlations \cite{donahue2013decaf, girshick2014rich}, can be instrumental in analyzing time-series data derived from vehicular sensors. Meanwhile, ViTs offer a complementary strength by adeptly handling image-based inputs, capturing intricate patterns and long-range dependencies that are often present in vehicular scenarios, such as V2X cooperative perception \cite{10638}. The synergy between these architectures can offer a more nuanced understanding of the environment, thereby enhancing the predictive accuracy of vehicular network management systems.

Moreover, the dynamic nature of vehicular networks, with constantly changing environments and vehicular movements, necessitates the use of models that can adapt over time. Gated Recurrent Units (GRUs) \cite{cho2014learning} emerge as a suitable choice in this context, given their capability to model temporal dependencies and retain relevant information across time steps. Integrating GRUs allows for the effective capture of the temporal evolution of vehicular network states, providing a robust framework for making informed predictions about potential blockages.

In light of these considerations, the motivation for integrating CNNs, ViTs, and GRUs in our approach is driven by the need to effectively utilize the rich multimodal data available in 6G-enabled vehicular networks. By harnessing the strengths of these advanced machine learning architectures, we aim to address the complex challenge of blockage prediction, ensuring seamless communication in the intelligent transportation systems of the future.

\subsection{Related Work}
The aforementioned challenges have motivated numerous researchers to propose innovative solutions for blockage detection and prediction to address the LoS blockage issue in V2X networks. These proposed solutions make use of various approaches as well as different types of data, such as visual data from cameras \cite{06255} and tabular data, including channel and beamforming vectors \cite{1259}.
Moreover, with the rapid evolution of artificial intelligence (AI) and machine learning techniques, researchers have employed advanced tools to predict LoS using different data sources.

Initially, the solutions were based on the multi-connectivity approach, such as the one proposed by \cite{7528494}, which introduced a centralized multi-cell solution to improve the quality of connections in the face of disconnections caused by LoS blockages in Heterogeneous Networks (HetNets). The solution is based on keeping track of the connection link between the user and multiple base stations, where the centralized unit collects information from all the base stations to make a decision about the quality of the connection.
However, these solutions detect the disconnections without anticipating them, which causes the user to get disconnected for a while until the base station recovers the connection link, causing communication latency.
Therefore, more information is needed to allow the base station to predict the NLoS state and the optimal beamforming codebook to be used before losing the connection. For this reason, different tabular data sources and types were explored to extract the needed information for the machine-learning model to anticipate blockage.

For instance, in \cite{8166758}, the authors used a series of channel state information (CSI) measured by signal transmissions in an indoor office environment with a Recurrent Neural Network (RNN) model consisting of a Long Short-Term Memory (LSTM) block \cite{6795963} to predict the connection state (LoS or NLoS). 
Also, the authors of \cite{9121328} proved the ability of neural networks to detect the current link status using the sub-6GHz channel from the DeepMIMO dataset script \cite{06435}, which generates a labeled sub-6GHz and mmWave channels set.
On the other hand, authors in \cite{02723}, provided a predictive solution to anticipate the stationary blockage for a single mmWave user using past observations of beamforming vectors, with an approach based on the GRU network \cite{1259}.
The mentioned approaches achieved outstanding performance in predicting stationary blockages. However, in many practical scenarios, such as vehicular networks, the movement of the environment and the users can cause blockages to occur and disappear rapidly. In such cases, predicting only stationary blockages may not be sufficient to ensure reliable and efficient communication. Therefore, it is important to also consider the prediction of dynamic or non-stationary blockages

Due to the limitations of previous solutions, the research community tried to explore other data types to reach better prediction performance for static and dynamic future blockages.
As a novel solution, the authors in \cite{06255} proposed a blockage prediction framework using a publicly available vision-aided wireless communications (ViWi) framework dataset \cite{06257}, which provides RGB images of different communication scenes and a tabular wireless data generator script. However, it considered only single-user communication settings.

Finally, at the time of writing this paper, the state-of-the-art solution was proposed by \cite{9512383}, where the authors consider both tabular and vision data to form a multimodal data approach for multi-user communication settings.
This approach is based on a centralized DL solution that takes advantage of the extracted information from both beam vectors and the images, first to detect the presence of a user, and then detect whether there is a possible future blockage or not using the position of the detected user.
This solution has shown an impressive performance improvement in terms of detecting future LoS blockages. However, it is extremely dependent on the performance of the object detection component which is sometimes inaccurate and impacts the overall performance.

The methodologies referenced in previous research aimed to optimize the utilization of wireless data to forecast future LoS blockages. Despite these efforts, the inclusion of alternative data sources, such as images, became necessary. However, the integration of multiple data modalities poses challenges in extracting relevant information. Therefore, a more robust feature extraction architecture is needed to effectively handle multimodal data and enable the extraction of valuable information for a more comprehensive understanding of the environment, leading to more precise predictions of future connection states.

\subsection{Contribution}
In this paper, we present a multimodal vision transformer-aided predictive framework that anticipates future connection states using both images collected using an RGB camera and beam vectors collected from the perspective of the base station.
Our approach focuses on developing an advanced AI-driven framework that addresses the challenges of anticipating blockages, including both static and dynamic obstacles. This framework efficiently leverages the rich information contained in multimodal data.
In the development of our vision-aided blockage prediction method, we have employed the ViWi dataset \cite{06257}, as it was the most comprehensive and pertinent dataset available at the time of conducting this research.

To establish a robust and efficient predictive framework, we introduce a unique time series DL architecture that incorporates a feature extraction component. Specifically, our framework utilizes a customized architecture of CNN and ViT to extract needed information from the input images and beam vectors. The proposed architecture allows the extraction of a comprehensive and detailed representation of the multimodal input data. This, in turn, produces more accurate and robust predictions compared to existing methods.
Specifically, this paper's contributions are summarized as follows:
\begin{itemize}
    \item Our approach employs a time-series DL model, specifically a GRU neural network, to predict future LoS connection states. What sets our approach apart is the quality of the features extracted by our unique feature extraction architecture. By leveraging advanced DL techniques such as CNNs and ViTs, our framework embeds multimodal data efficiently and provides a substantial volume of pertinent information for the learning process.
    \item Our proposed approach goes a step further by reducing dependence on the object detection component commonly used in vision-aided predictive frameworks. By doing so, we create a more resilient predictive framework that is less susceptible to potential object detection failures or misinterpretations. This reduction in dependency results in a higher degree of consistency and reliability in predictions, ensuring that the system performs well even in scenarios with challenging visual conditions or limitations in the computer vision component's performance.
    Through experiments and comparative analysis, we demonstrate that our solution leads to significantly improved prediction accuracy and stability compared to the state-of-the-art solution.
    
\end{itemize}

\subsection{Organization}
This paper is organized as follows. In Section II, we describe the system model considered in our study. In Section III, we formulate the V2X blockage prediction problem and provide a detailed analysis of the challenges. Section IV presents our proposed methodology, which includes feature extraction using ViT and classification using a GRU neural network. In Section V, we provide details on the implementation of the proposed solution and the obtained numerical results, including a comparison with the state-of-the-art works. Finally, in Section VI, we present our conclusion and highlight directions for future work.

\section{System and Channel Models}
In this section, we provide a detailed description of the system and channel models used for our proposed V2X blockage prediction methodology.

\subsection{System Model}
\begin{figure}[!t]
    \centerline{\includegraphics[width=\columnwidth]{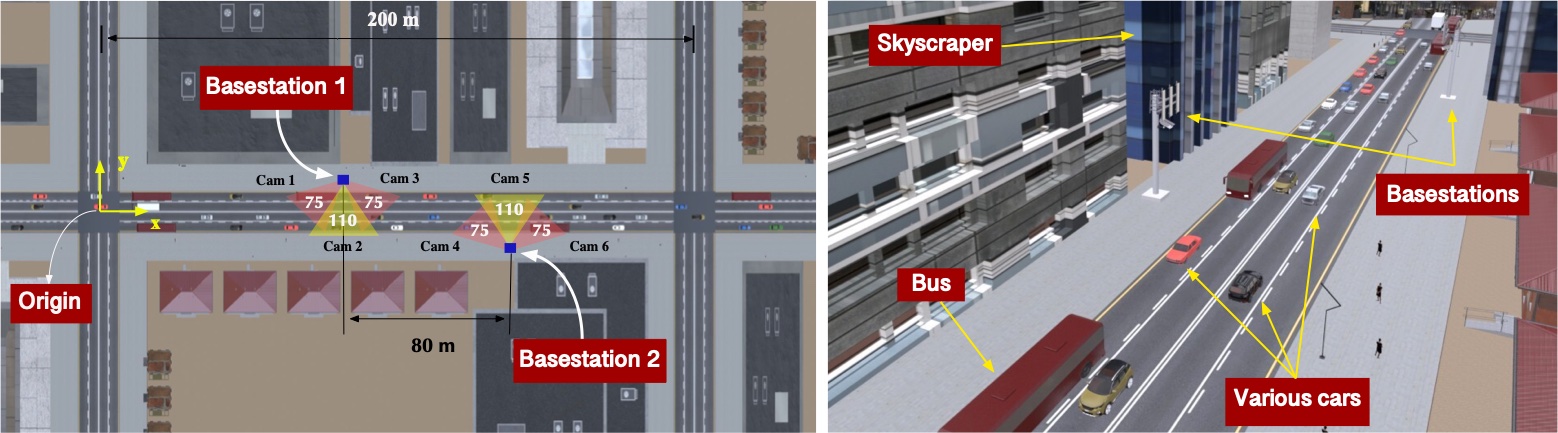}}
    \caption{A top-view and a perspective view of the ViWi outdoor scenario \cite{06257}.}
    \label{fig7}
\end{figure}

In a vehicle network, the environment is formed by dynamic users and dynamic objects that could be detected as possible obstacles, as presented in Figure \ref{fig7}, with a base station equipped with a standard-resolution RGB camera.

We consider a highly dynamic vehicular network environment, where base stations use the beamforming codebook technique, which is helpful to target the desired user directly without losing signal energy in undesired directions. 
Beamforming vectors can provide information about the direction of the user, this provides the base station with 3D awareness, which is significantly helpful for both detecting the existence of the user, as well as tracking its location.
Moreover, images serve as a data type rich with valuable information about the environment. Visual data is highly informative, offering essential details about the user, including their position, distance, velocity, size, and more.

Formally, at a time step $t^{\prime}$ in a time interval of $p$ steps and for a given user $u$, an observation $\mathcal{X}_{u}$ is defined as follows:
\begin{equation}
\mathcal{X}_{u}[t^{\prime}]=\left\{\left(\mathbf{A}_{u}[t], b_{u}[t]\right)\right\}_{t=t^{\prime}-p+1}^{t^{\prime}}.
\end{equation}
where $\mathbf{A}_{u}$ is a series of image frames of the environment taken from the perspective of the base station, and $b_{u}$ is the beam index of the corresponding optimal beamforming vector from the predefined beam codebook $\mathcal{F}$ while communicating with user $u$.

Our vehicular network consists of an outdoor environment including small-cell mmWave base stations using beamforming.
The decision about the optimal beam vector index while communicating with a user $u$ is extracted from a predefined beamforming codebook $\mathcal{F}$:
\begin{equation}
\mathcal{F}=\left\{\mathbf{w}_{n}\right\}_{n=1}^{N},
\label{eq3}
\end{equation}
where $N$ denotes the number of the beamforming vectors present in the codebook and $\mathbf{w}_{n} \in \mathbb{C}^{M \times 1}$ is given by:
\begin{equation}
\mathbf{w}_{n}=\frac{1}{\sqrt{M}}\left[1, e^{j \frac{2 \pi}{\lambda} d \sin \left(\phi_{n}\right)}, \ldots, e^{j(M-1) \frac{2 \pi}{\lambda} d \sin \left(\phi_{n}\right)}\right]^{T},
\label{eq4}
\end{equation}
where $\phi_{n} \in\left\{\frac{2 \pi n}{N}\right\}_{n=0}^{N-1}$ defines the uniform quantization of the azimuth angle and $\lambda$ as the wavelength.
In this system, we consider $M$ elements uniform linear array (ULA) antennas at the base stations, which is equipped with three RGB cameras targeting the frontal, right, and left directions.

\subsection{Channel Model}
In our research, the experimental results are derived from data samples obtained using the ViWi data generation framework, which employs the ray-tracing software Wireless InSite to simulate realistic mmWave channel conditions.
The channel model adopted throughout this paper is a geometric mmWave channel model with $L$ clusters. The communication system utilizes orthogonal frequency division multiplexing (OFDM) with $K$ subcarriers and a cyclic prefix of length $D$.
For each user $u$, we define the downlink channel with received signal $y_{u, k}$ as follows

\begin{equation}
y_{u, k}=\mathbf{h}_{u, k}^{T} \mathbf{w}_{n^{*}} x+n_{k},
\label{eq5}
\end{equation}
where $\mathbf{h}_{u, k} \in \mathbb{C}^{M \times 1}$ denotes the channel between the base station and user $u$ at sub-carrier $k$, 
and $\mathbf{w}_{n^{*}}$ is the optimal beam vector that maximizes the received SNR at the receiver selected from the predefined beam codebook $\mathcal{F}$, and the noise sample $n_{k}$ follows a complex Gaussian distribution $\mathcal{N}_{\mathbb{C}}\left(0, \sigma^{2}\right)$, given by \cite{9512383}

\begin{equation}
\mathbf{h}_{u, k}=\sum_{d=0}^{D-1} \sum_{\ell=1}^{L} \alpha_{\ell} e^{-\mathrm{j} \frac{2 \pi k }{K} d} p\left(d T_{\mathrm{S}}-\tau_{\ell}\right) \mathbf{a}\left(\theta_{\ell}, \phi_{\ell}\right),
\end{equation}
where $\alpha_{\ell}$ is the path gain including path loss, $p$ is the pulse shaping filter, $\tau_{\ell}$ is the delay, $T_{S}$ is the sampling period, and $\mathbf{a}$, $\theta_{\ell}$, $\phi_{\ell}$ are the array manifold vector,  the azimuth and elevation angles of arrival.

\section{Problem Formulation}
In this section, we formally define the problem of V2X blockage prediction, which is the main focus of this paper.
Let us consider a highly dynamic 6G vehicular network where vehicles are communicating with a base station (BS) through a mmWave channel. We assume directional communication with the aid of beamforming techniques where the vehicles and BS are equipped with a beamforming codebook containing a set of predefined beam vectors, and they adapt the beam direction according to the CSI to optimize the communication performance. The goal of the beamforming algorithm is to maximize the signal-to-interference-plus-noise ratio (SINR) by steering the main beam toward the intended receiver.
Our goal is to predict the occurrence of these blockages and their duration in advance, hence allowing the vehicles and the BS to switch to a more suitable beam direction to maintain the communication link.

Considering a future interval in which we predict the blockage.
Let $f$ be the size of the future interval, and let $l_{u}[t] \in\{0,1\}$ be the LoS connection state indicator for the upcoming time step $t$, which is equal to 0 if an LoS is available and 1 if not (NLoS: blockage detected), and $\mathcal{L}_{u}[t^{\prime}]=\left\{l_{u}[t]\right\}_{t=t^{\prime}+1}^{t^{\prime}+f}$ is the set of connection statuses in the future interval.
The time step referred to aligns with the data generation framework of the ViWi dataset, where a 'time instance' does not specify an exact duration in conventional units like seconds or minutes. It reflects the duration used to generate each data point in the dataset.

We define the global future link status $s_{u}$ as a binary indicator of whether we have an LoS blockage in the future $f$ time steps, where:
\begin{equation}
s_{u}[t^{\prime}]= \begin{cases}
0, & \text{if } l_{u}[t]=0, \forall t \in\left\{t^{\prime}+1, \ldots, t^{\prime}+f\right\} \\ 
1, & \text{otherwise}
\end{cases},
\end{equation}
which is equal to 0 if a LoS connection will be maintained for the next $f$ time steps and 1 if not (a blockage is predicted to occur in the future interval).
The principal objective is to design a DL framework capable of extracting the needed information from the multimodal data to maximize the prediction accuracy of future connection states $\hat{s}_{u}$ for all users $u$ in the set of users $U$:
\begin{equation}
\max \prod_{u=1}^{U} \mathbb{P}\left(\hat{s}_{u}=s_{u} \mid \mathcal{X}_{u}\right).
\end{equation}

In the next section, we provide a description of the methodology of the proposed solution.

\section{Proposed Solution Methodology}
\begin{figure*}[h!]
    \centering
    \includegraphics[width=1.9\columnwidth]{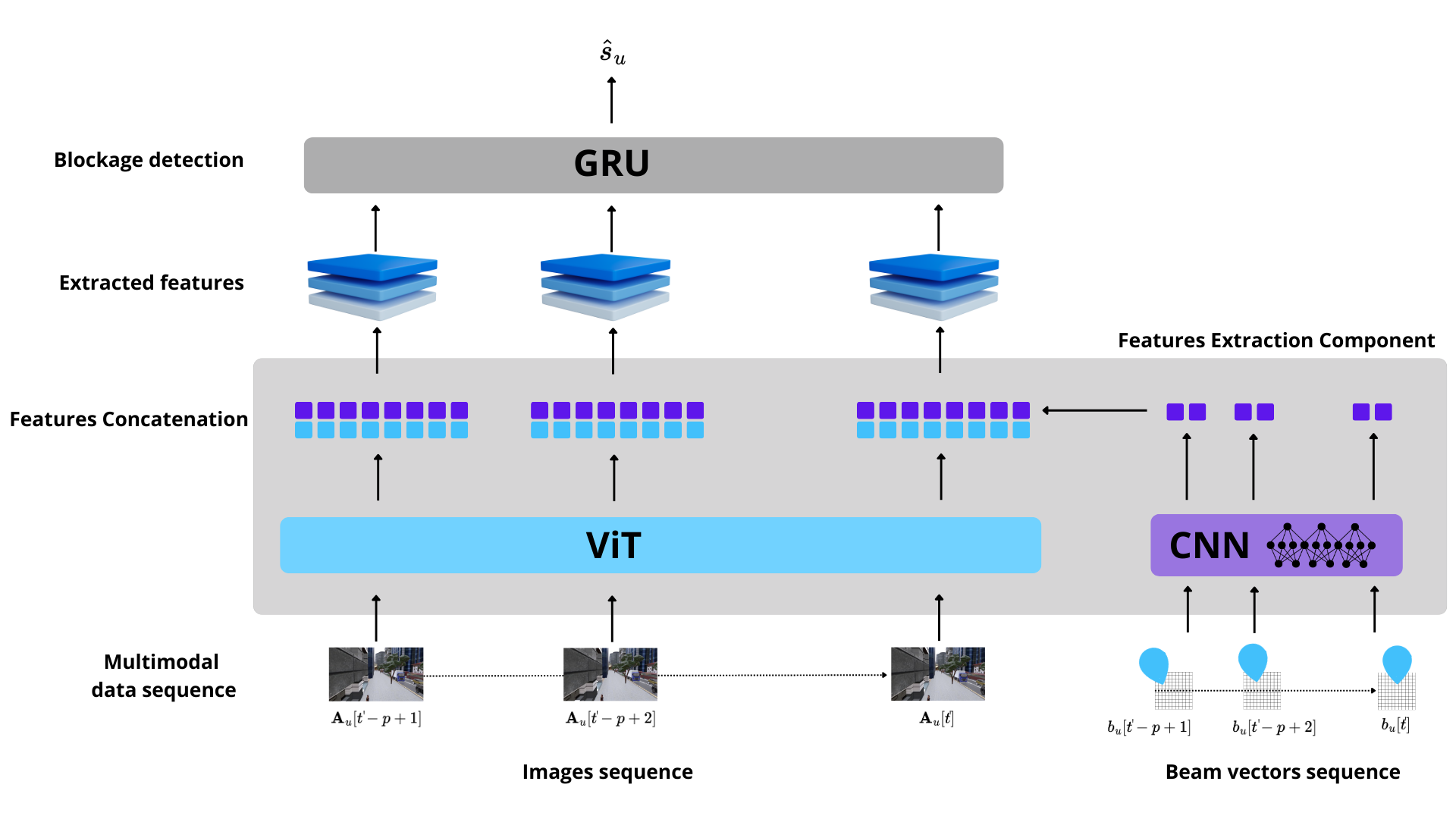}
    \caption{Proposed V2X LoS ViT-CNN framework.}%
    \label{framework}
\end{figure*}
The proposed methodology described in Figure \ref{framework} aims to predict future obstacles in LoS vehicular communications between base stations and vehicles using multimodal data that includes images and beamforming vectors.
This task is becoming increasingly complex due to the higher frequency bands that will be used in future networks, the high mobility of users, and the heterogeneity of the used data to make the decision.
 
There are mainly two challenges that need to be addressed to realize the full potential of this methodology. 
The first one is related to the type of data we are dealing with. 
In fact, since the data comes from different modalities, such as images and beamforming vectors, there may be differences in the data format, size, resolution, and type of information that could be extracted from each data type. This can make it difficult to integrate the data and extract meaningful features. The other challenge, proven by the available state-of-the-art solution \cite{9512383}, is the high dependency on the object detection component, adversely affecting the overall performance with any small errors caused by the object detection model, which can limit the scalability of the methodology. 

To address these challenges effectively, we propose an efficient feature extraction technique capable of handling the heterogeneity of the data and extracting essential information without relying on an object detection component. Additionally, we optimize the training process of the feature extraction component to accommodate different data types and utilize it as an embedding component.

\begin{figure}[!t]
    \centering
    \includegraphics[width=1 \columnwidth]{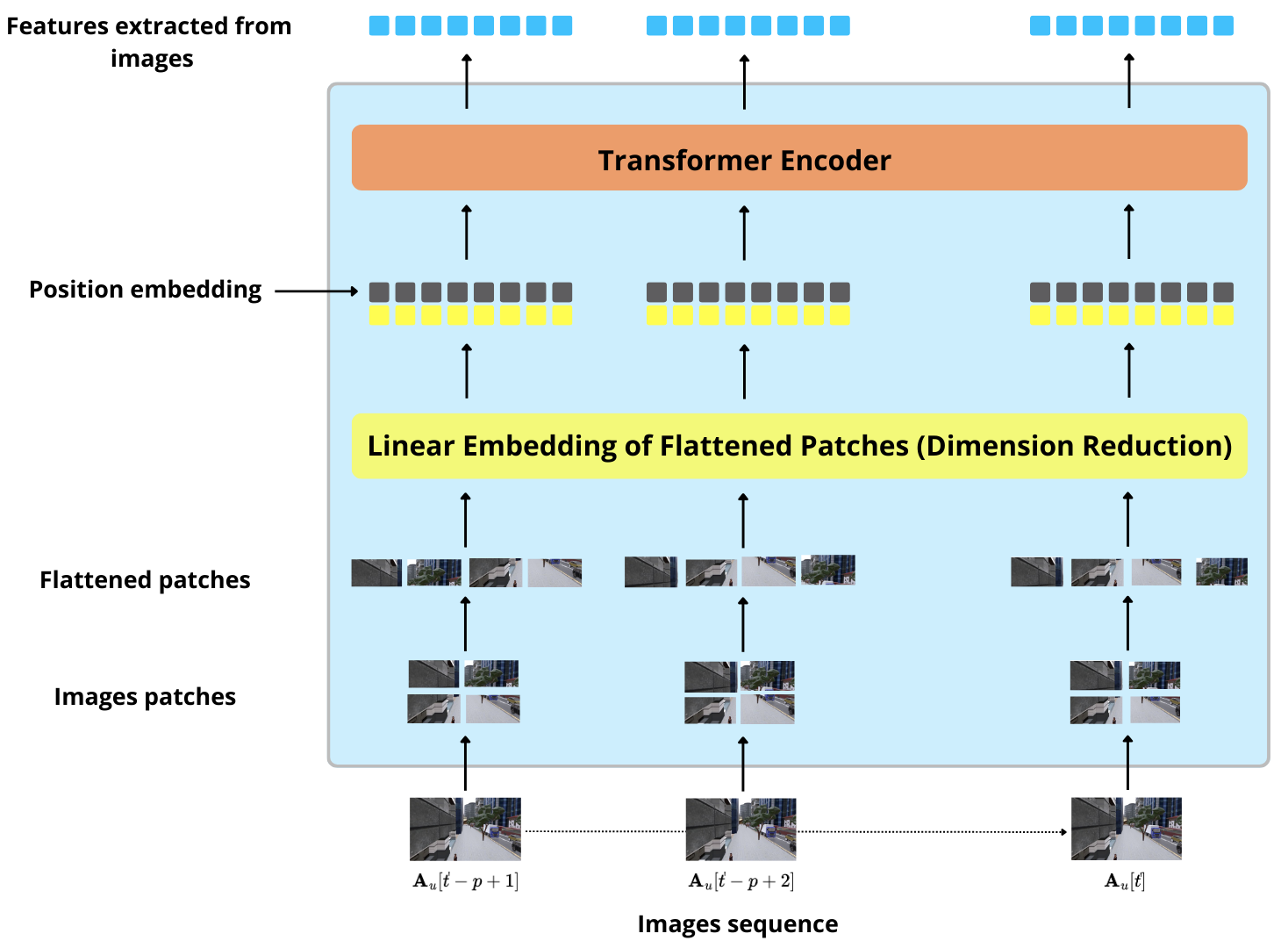}
    \caption{The architecture of our proposed ViT model.}
    \label{vit}
\end{figure}

\begin{figure}[!t]
    \centering
    \includegraphics[width=.3 \columnwidth]{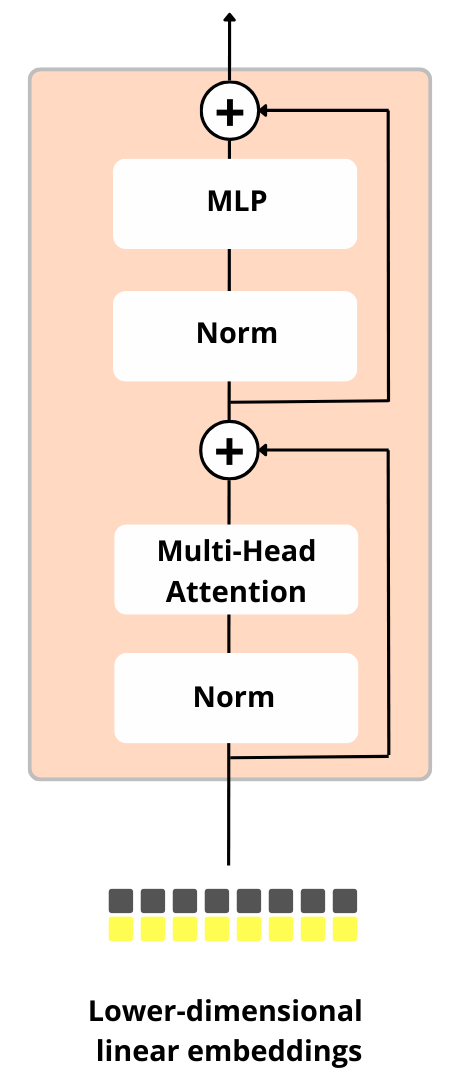}
    \caption{The illustration of the transformer encoder inspired by\cite{vaswani2017attention}.}
    \label{enc}
\end{figure}

\subsection{Feature Extraction}
To achieve efficient information extraction essential for the final decision-making process, we propose a novel transformer-based architecture for feature extraction from multimodal data for blockage detection in V2X, as shown in Figure \ref{framework}.
In this design, we apply CNNs for processing tabular wireless data and use ViTs to handle images. This hybrid approach takes advantage of the unique strengths of both architectures, facilitating effective information capture and extraction from diverse modalities.
The feature extraction component is designed to extract useful information $\mathcal{I}_{u}$ from the input data $\mathcal{X}_{u}$, which consists of beamforming codebook vectors and camera images $\left(\mathbf{A}_{u}[t], b_{u}[t]\right)$. 
The proposed architecture consists of two parallel branches: one branch for CNN-based feature extraction and another branch for ViT-based feature extraction, then merged to provide the extracted information to the final predictive element.
These two branches will be described in detail in the following subsections.

\begin{table}[h]
\centering
\caption{Architecture Parameters}
\label{architecture}
\resizebox{\columnwidth}{!}{
\begin{tabular}{|c|c|}
\hline
\textbf{Parameter} & \textbf{Value} \\ \hline

\multicolumn{2}{c}{\textbf{ViT Configurations}} \\ \hline
Number of Transformer Layers & 6 \\ \hline
Hidden Units in Each Transformer Layer & 2048 \\ \hline
Number of Attention Heads & 8 \\ \hline
Input Image Size & [224x224] \\ \hline
Patch Size & [16x16] \\ \hline
Embedding Dimension & 512 \\ \hline

\multicolumn{2}{c}{\textbf{CNN Architecture Parameters}} \\ \hline
Number of Convolutional Layers & 4 \\ \hline
Number of Filters in Each Convolutional Layer & [32, 64, 128, 256] \\ \hline
Filter Sizes & [3x3, 3x3, 3x3, 3x3] \\ \hline
Stride & [1, 1, 1, 1] \\ \hline
Activation Function & ReLU \\ \hline

\multicolumn{2}{c}{\textbf{GRU Architecture Parameters}} \\ \hline
Number of GRU Layers & 2 \\ \hline
Hidden Units in Each GRU Layer & [256, 128] \\ \hline
Bidirectional GRU & No \\ \hline
Activation Function & Hyperbolic Tangent (tanh) \\ \hline
Dropout Probability & 0.3 \\ \hline
Output Layer & Fully Connected Layer \\ \hline
Output Units & 1 (Binary Classification) \\ \hline
Activation Function for Output & Sigmoid \\ \hline
Loss Function & Binary Cross-Entropy \\ \hline

\end{tabular}}
\end{table}

\subsubsection{CNN Branch}
The CNN branch focuses on processing tabular wireless data by extracting essential features from beamforming vectors $\left\{b_{u}[t]\right\}_{t=t^{\prime}-p+1}^{t^{\prime}}$. Our CNN architecture is designed to capture intricate spatial dependencies within the beam vectors. As shown in Table \ref {architecture}, it consists of 4 convolutional layers, with each layer equipped with an increasing number of filters, ranging from 32 to 256. A 3x3 filter size is employed consistently across all layers, ensuring an efficient receptive field. A stride of 1 is set for each convolutional layer, ensuring that the convolution operation scans the entire input. Rectified Linear Unit (ReLU) activation functions are applied after each convolutional layer, introducing non-linearity into the model.
The output of the CNN branch is a set of high-dimensional feature maps that capture essential patterns and relationships within the beamforming vectors.

\subsubsection{ViT Branch}
In our framework, the ViT branch is responsible for processing the images $\left\{\mathbf{A}_{u}[t]\right\}_{t=t^{\prime}-p+1}^{t^{\prime}}$, which provides visual information relevant to blockage detection. 
ViT is a state-of-the-art architecture for image understanding tasks, particularly for V2X information extraction applications \cite{10638}. 

In the ViT branch, as described in Figure \ref{vit}, the input images are divided into non-overlapping patches of size [16x16], which are then linearly embedded into a lower-dimensional representation of size 512 to obtain a sequence of patch embeddings. The patch size is optimized and set to be large enough to capture global patterns in the image.
A positional encoding is then added to the patch embeddings to provide the network with the positional information of each patch in the image.
These embeddings, along with position embeddings, serve as the input to the transformer encoder proposed in \cite{vaswani2017attention} described in Figure \ref{enc}.
The transformer encoder consists of multiple layers of self-attention mechanisms (MSA) and fully connected multi-layer neural network (MLP) blocks. Self-attention allows the model to attend to different parts of the image while capturing global relationships between patches.
The output of the ViT branch is a sequence of high-level embeddings that encode the spatial information and relationships within the images. These embeddings effectively capture the visual characteristics that contribute to blockage detection.

\subsection{Time Series Prediction}
Time series forecasting has traditionally been dominated by linear and ensemble methods. These methods are highly interpretable and efficient for a variety of problems, particularly when coupled with feature engineering.
However, with the advent of RNN in the 1980s, followed by more advanced RNN structures, such as LSTM in 1997 \cite{6795963}, and more recently, GRU in 2014 \cite{1259}, DL techniques have enabled the learning of complex relationships between sequential inputs and outputs with limited feature engineering. These RNN techniques have enormous potential for analyzing large-scale time series in previously unfeasible ways.

GRUs, a newer version of RNNs, show similar performances as RNNs and LSTMs while being significantly faster to compute. GRUs record long-term dependencies without any cell state by using reset and update gates. The update gate identifies how much past information must be kept, while the reset gate determines how much past information must be ignored. GRUs are often faster and require less memory than LSTMs since they require fewer tensor operations.

In our methodology, after extracting relevant features using our customized architecture, the next step is to pass these features through a time series computer vision classification component to make a binary decision.

For this purpose, we use a GRU  as the temporal modeling component of our framework. Our GRU network consists of 2 stacked layers, each layer comprising 256 and 128 hidden units, respectively. The GRU architecture operates in a unidirectional manner, as bidirectional processing is deemed unnecessary for the given task due to the inherently causal nature of the LoS prediction task, where future information cannot influence past events.
Hyperbolic Tangent (tanh) activation functions are applied, and a dropout probability of 0.3 is introduced to mitigate overfitting. The final layer consists of a fully connected layer with a single neuron, for the final binary classification, and the sigmoid activation function ensures that predictions fall within the [0, 1] range. Binary Cross-Entropy is chosen as the loss function, making it well-suited for our binary LoS prediction task. The GRU network serves as the decision-maker in our multimodal LoS prediction system, synthesizing both spatial and temporal information to predict future LoS states accurately.

\section{Implementation and Numerical Results}
In this section, we elaborate on the vehicular communication environment that was used for data generation, as well as the process followed to prepare the data used in the proposed framework. We also describe the training of the feature extraction components and the time series computer vision classifier.
Additionally, we provide numerical results and a detailed evaluation of the performance of the proposed methodology in order to evaluate its effectiveness in predicting future obstacles in vehicular communications.

\begin{table}[h]
\centering
\caption{Implementation Parameters}
\label{implementation}
\resizebox{\columnwidth}{!}{
\begin{tabular}{|c|c|}
\hline
\textbf{Parameter} & \textbf{Value} \\ \hline
\multicolumn{2}{c}{\textbf{Data Generation Parameters}} \\ \hline
Number of base stations & $2$ \\ \hline
Number of antennas per base station & $128$ \\ \hline
Beam codebook size & $128$ \\ \hline
Antenna spacing & $0.5 \lambda$ \\ \hline
bandwidth & $0.2$ GHz \\ \hline
Number of OFDM subcarriers & $128$ \\ \hline
OFDM sampling factor & $1$ \\ \hline
Number of users & $60$ \\ \hline
Number of scenes (number of frames) & $1000$ \\ \hline

\multicolumn{2}{c}{\textbf{ViT Configurations}} \\ \hline
Pre-trained Model & Yes (pretrained on ImageNet \cite{5206848}) \\ \hline
Learning Rate for Fine-Tuning & $1e-5$ \\ \hline
Batch Size for Fine-Tuning & $32$ \\ \hline

\multicolumn{2}{c}{\textbf{CNN Training Parameters}} \\ \hline
learning rate & $1e-3$ \\ \hline
batch size & $64$ \\ \hline
drop out probability & $0.2$ \\ \hline
number of epochs & $1000$ \\ \hline
Optimizer & adam optimizer \\ \hline

\multicolumn{2}{c}{\textbf{GRU Training Parameters}} \\ \hline
Number of Training Epochs & $1000$ \\ \hline
Learning Rate & $1e-4$ \\ \hline
Batch Size & $32$ \\ \hline
Padding for Variable-Length Sequences & Zero Padding \\ \hline
Gradient Clipping & Yes (Threshold: $1.0$) \\ \hline
Input sequence: $p$ & $8$ \\ \hline
Output sequence $f$ & $3$ \\ \hline
Optimizer & adam optimizer \\ \hline

\end{tabular}}
\end{table}

\subsection{Environment and Data Setup}
In this section, we illustrate the framework used to build the needed environment and generate the wireless and image datasets. 
Specifically, we start with the definition of the scenarios from the ViWi dataset generator \cite{06257}, 
followed by the detailed process for raw data generation,
and finally, the data processing is applied to our raw data to provide the needed data sequences for the DL time series model.

\subsubsection{Environment Preparation}
Our environment is built using the "ASUDT1" scenario from ViWi open-source dataset \cite{06257}, which contains an outdoor multi-user environment. 
As illustrated in Figure \ref{fig7}, the scenario illustrates a busy downtown street, along with its various elements, such as cars, buses, trucks, skyscrapers, buildings, lamp posts, etc. 
The experimental setup consists of a pair of base stations with uniform linear array mmWave antennas. The base stations are placed 80 meters apart, facing opposite directions, and operating at a frequency of 28 GHz. A custom Matlab script is utilized to optimize the size and shape of the antenna array. 

In addition, each base station is equipped with three cameras, which are strategically placed to capture the surrounding environment from different angles. The cameras are labeled from 1 to 6 and have overlapping fields of view. Specifically, cameras 3 and 4, situated on the left side of the first base station and the right side of the second base station, respectively, share almost identical fields of view. 
The use of multiple cameras with overlapping views enables us to capture different perspectives of the environment and ensures a more comprehensive and accurate dataset for our predictive model.
The ViWi data generation script enables the manipulation of the number of users and the number of scenes that the scenario has, where a scene is one frame in a video captured by the RGB camera.
Each vehicle, equipped with a mmWave radio receiver, continuously moves in one direction in one of the four road lines with a variable velocity. 

\subsubsection{Data Generation Pipeline}
\begin{figure}[!t]
    \centerline{\includegraphics[width=\columnwidth]{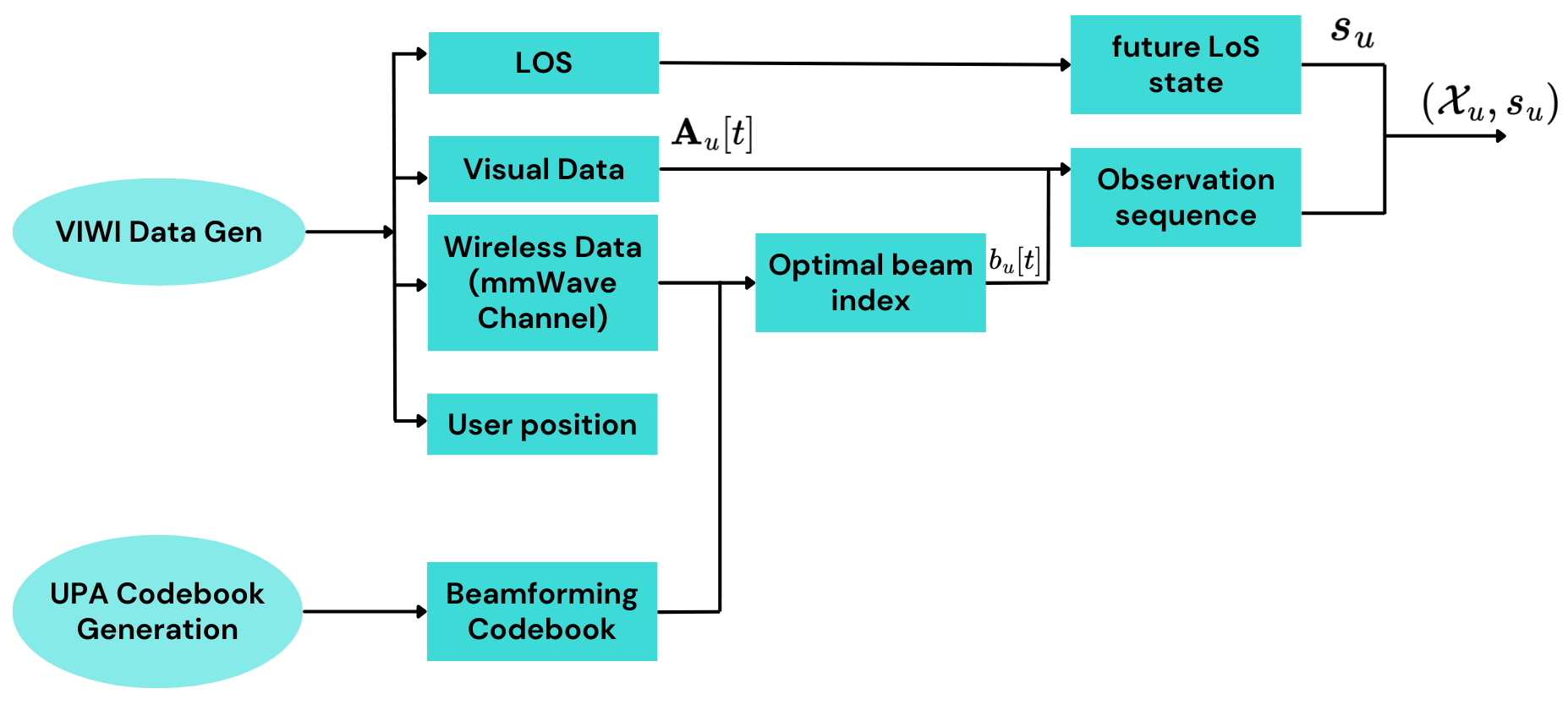}}
    \caption{Data generation pipeline.}
    \label{fig8}
\end{figure}
As depicted in Figure \ref{fig8}, the data generation pipeline for this project begins with the ViWi raw data generator, which utilizes the parameters provided in Table \ref{implementation}. The ViWi generator produces a 4-tuple data point, which includes an image of a selected scene, an mmWave channel for each user for each OFDM subcarrier for each of the $M$ antennas at time step $t$, the coordinates of the user position in the visual field of the selected base station, and the connection state indicating whether there is a LoS connection or not.

The beam codebook vectors are then obtained using a uniform planar array (UPA) codebook generation Matlab script, which generates a beam-steering codebook design. Using the generated channels, we select the beam vector that maximizes the received SNR at the receiver from the codebook, resulting in the following optimal beam vector for a total of K subcarriers.

\begin{equation}
\mathbf{w}^{\star}=\underset{\mathbf{w} \in \mathcal{F}}{\operatorname{argmax}} \frac{1}{K} \sum_{k=1}^K \mathbb{E}\left[\left\|\left(\mathbf{h}_k\right)^T \mathbf{w}\right\|_2^2\right],
\end{equation}

Finally, using the LoS state of the future $f$ steps, the pipeline constructs the target of the data and embeds the resulting dataset in the form of observation and target. The observation is a pair of sequences of images and beam vectors $\left(\mathbf{A}_{u}[t], b_{u}[t]\right)$, while the target is the future LoS state $s_{u}$. This comprehensive pipeline generates a large and diverse dataset to train and test the proposed solution for V2X blockage prediction.

\subsubsection{Data Processing}
To render the data suitable for a time series DL model, we need to form a dataset composed of sequences. 
Each data point consists of $p$ successive pairs of images and beam vectors $\left(\mathbf{A}_{u}[t], b_{u}[t]\right)$ for $t \in\left\{t^{\prime}-p+1, \ldots, t^{\prime}\right\}$, and the label $s_{u}$ which is a binary state indicating whether we have a blockage in the next $f$ time steps.
Thus, we have a sequence of $p$ pairs with information about the past $p$ time steps, and the target $s_{u}$ presenting the information about the future $f$ time steps.
In our experiments, we set $p$ to 8 and $f$ to 3.

After creating feature columns, such as time-lagged observations, making roughly $7000$ data points, we divide the dataset into three parts: training, validation, and test sets. It is noted that since our data is time-dependent, it is crucial to preserve the temporal sequence. Therefore, no shuffling has been applied to our dataset.

\begin{figure}[!h]
\centering
\subfloat[Training]{%
  \includegraphics[width=1\linewidth]{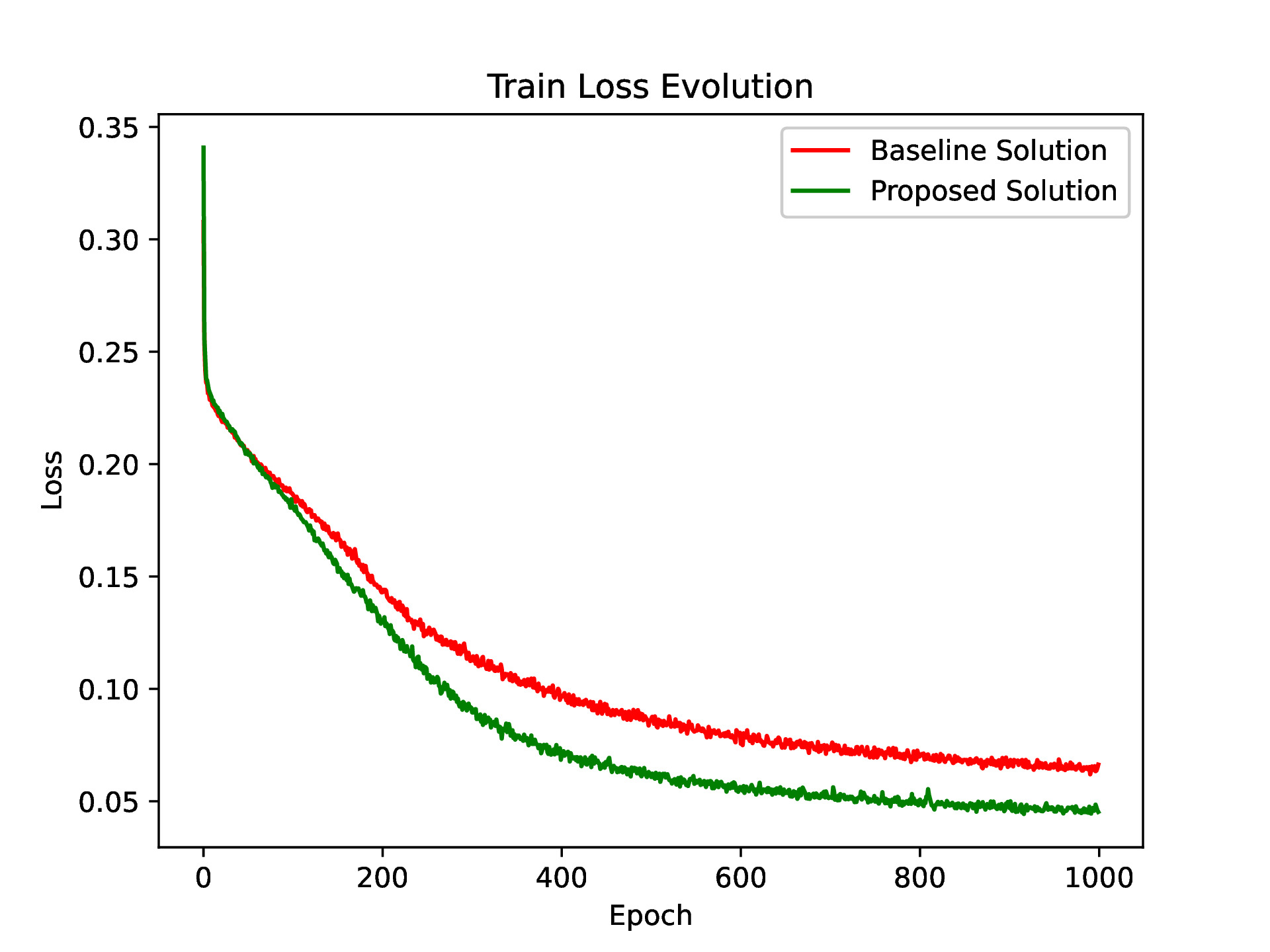}%
}
\hfil
\subfloat[Validation]{%
  \includegraphics[width=1\linewidth]{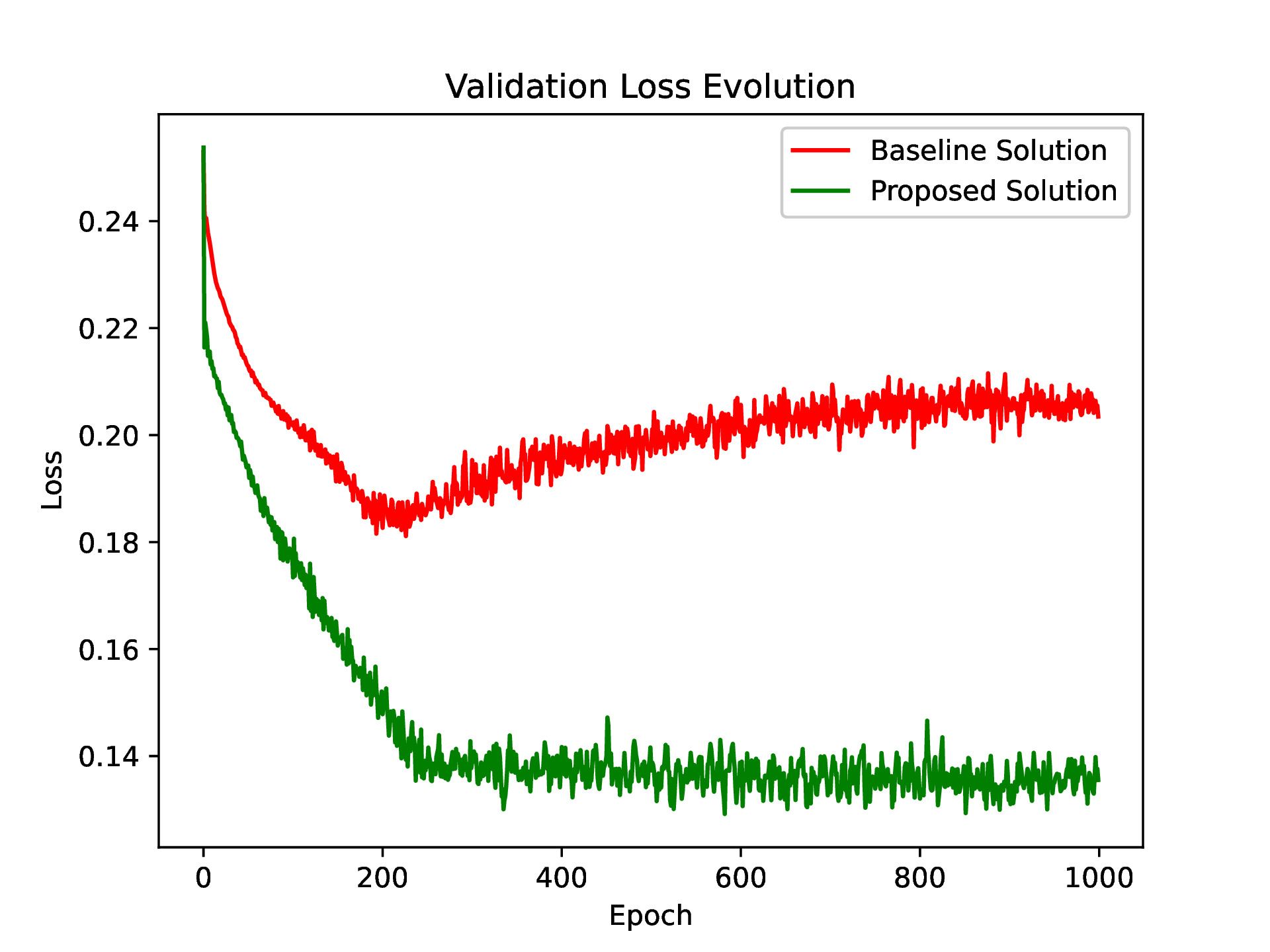}%
}
\caption{Evolution of loss function while training the proposed solution and the baseline method over both (a) training and (b) validation datasets.}
\label{loss}
\end{figure}

\begin{figure}[!h]
\centering
\subfloat[Training]{%
  \includegraphics[width=1\linewidth]{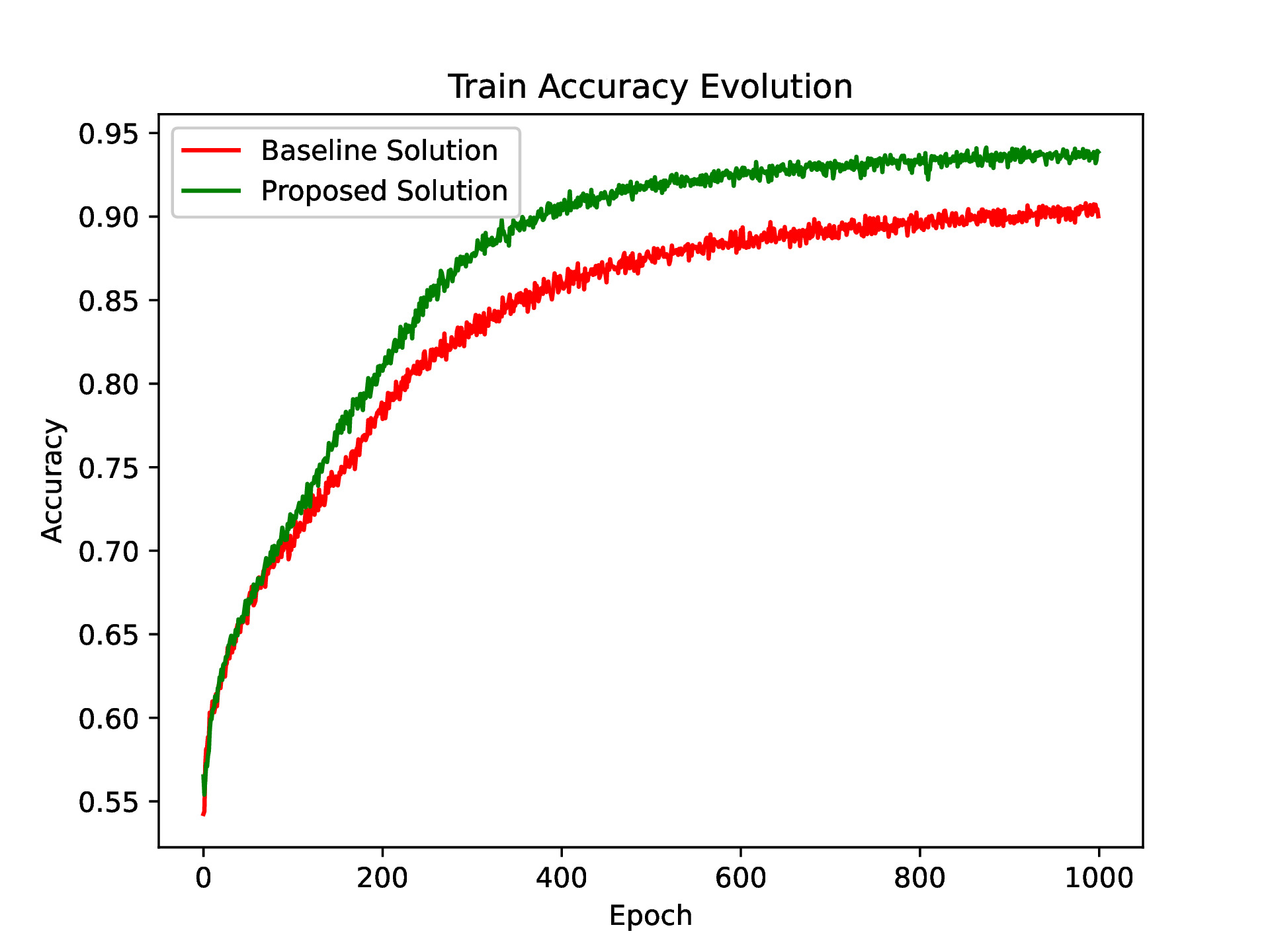}%
}
\hfil
\subfloat[Validation]{%
  \includegraphics[width=1\linewidth]{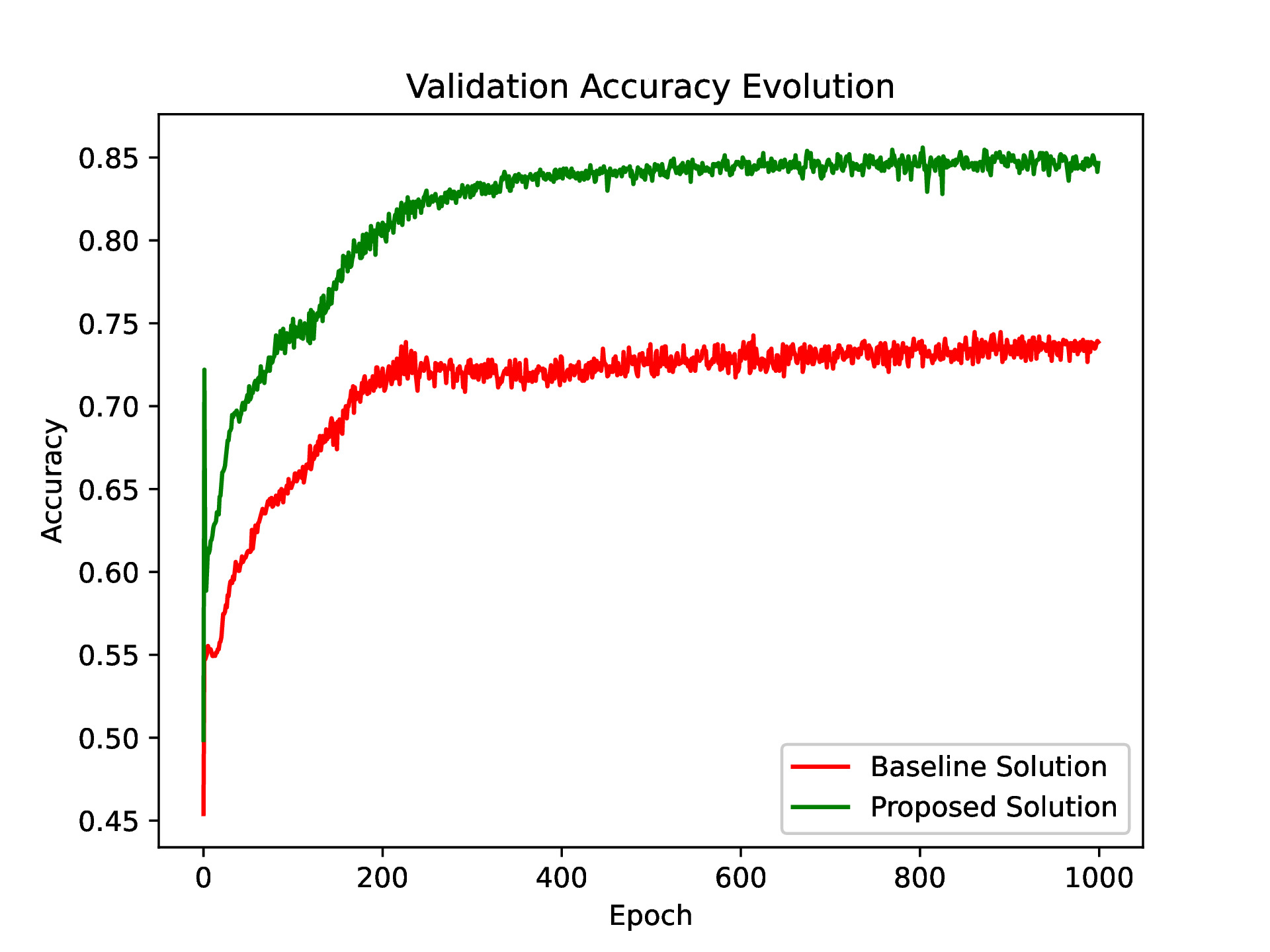}%
}
\caption{Evolution of accuracy while training the proposed solution and the baseline method over both (a) training and (b) validation datasets.}
\label{acc}
\end{figure}

\subsection{Implementation}
\subsubsection{Features Extraction}
The feature extraction component of our multimodal LoS prediction framework plays a crucial role in processing the multimodal data, which includes both images and beam vectors. We adopt a two-branch approach, leveraging the power of both ViT and CNN architectures for extracting relevant spatial and structural features. For the ViT branch, we employ a pre-trained model that has been fine-tuned on the ImageNet dataset \cite{5206848}. During fine-tuning, as shown in Table \ref{implementation}, we set the learning rate to 1e-5 and utilized a batch size of 32 to efficiently adapt the model to our specific LoS prediction task.

In parallel, the CNN branch focuses on extracting features from the beam vectors. Training parameters for the CNN include a learning rate of 1e-3, a batch size of 64, and a dropout probability of 0.2 to prevent overfitting. We train the CNN for 1000 epochs, allowing it to learn complex spatial dependencies within the beamforming data effectively. The Adam optimizer is employed for efficient weight updates during training. These two branches operate synergistically to provide a rich set of spatial and structural features, ensuring that the multimodal data is comprehensively processed for subsequent LoS state prediction.

\subsubsection{Detection}
In our framework, we employ a baseline GRU framework to detect blockages in the future $f$ time slots using the extracted information from the prepared sequences of past information during the last $p$ time slots. We trained the model using the parameters listed in Table \ref{implementation}, and the results are displayed in the following section.

As a baseline solution, we implement the object detection approach proposed in \cite{9512383}. This method proposes an object detection component to detect possible obstacles in the environment by extracting the coordinates of the bounding boxes of the detected objects and using only these coordinates with the beamforming vectors in the training process of the GRU model.
For the implementation of this solution, we used the latest object detection You Only Look Once (YOLO) \cite{02640} model: YOLOv7 \cite{02696}, followed by the same GRU model considered in our solution.

\subsection{Numerical Results and Evaluation}
\subsubsection{Comparison to State-of-the-art}
Using the described experimental setup, we trained and tested the proposed framework on the generated multimodal time-sequence data, and performed a thorough analysis of the results.
We compared our solution with the state-of-the-art approach described in \cite{9512383} that uses an object detection component as a baseline solution.

During the training process, we monitored the evolution of the training and validation loss values and the accuracy, which are common metrics for evaluating the performance of machine learning models. 
As displayed in Figures \ref{loss} and \ref{acc}, the results show a steady decrease in the loss value and an increase in the accuracy for both the training and validation sets over the course of training.
This behavior indicates that the model is learning the features and the patterns in the data effectively to predict the future LoS state. Moreover, our solution outperforms the state-of-the-art method where a faster decrease in the loss value and a faster increase in accuracy for both training and validation sets are clearly observed.
In fact, after $400$ epochs of training the GRU model, the proposed solution reaches a training accuracy of $0.905$  while the object detection solution is at $0.860$. Meanwhile, the validation accuracy of the GRU model reaches $0.839$ while that of the baseline is at 0.729.

The improved performance of our proposed solution on the validation dataset, which was not used during training, demonstrates its ability to generalize to new and unseen data. This is a crucial characteristic for a model to be practically applicable in real-world scenarios where the data is not perfectly clean and there may be variations in the input features.

\begin{figure}[!t]
    \centerline{\includegraphics[width=\columnwidth]{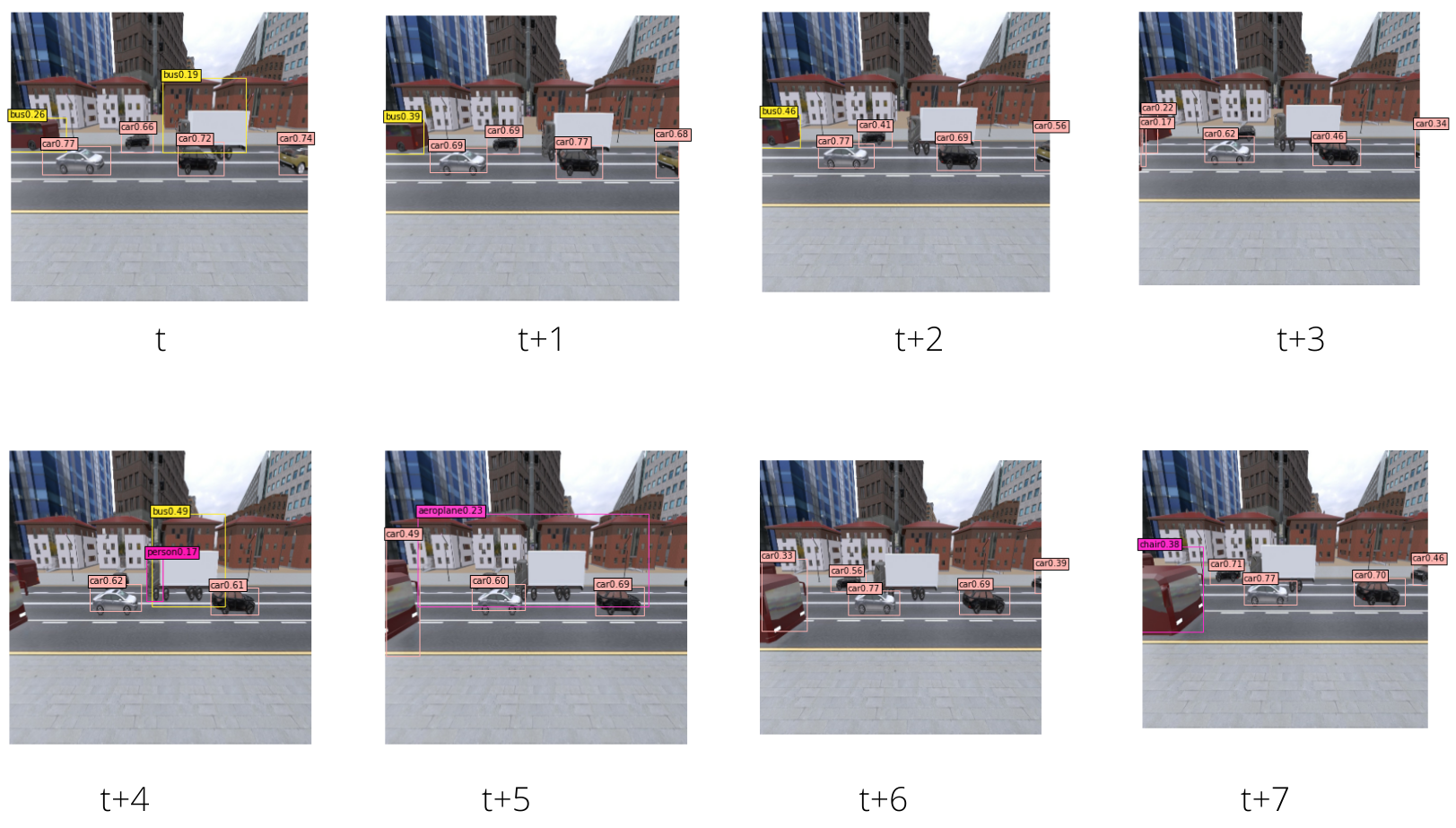}}
    \caption{Example of errors causing the limitations of the baseline solution.}
    \label{frames}
\end{figure}

Furthermore, the significant improvement in accuracy over the baseline solution proves the robustness of our solution. The baseline solution relies on the coordinates of the bounding boxes of detected objects, which makes it heavily dependent on the performance of the object detection model. As an example, Figure \ref{frames} shows two types of anomalies caused by the object detection component first the miss detection of the boundaries of the vehicles (the bus in time steps $t$, $t+4$, and $t+5$), second the non-detection of the vehicles (in time steps $t+1$, $t+2$, $t+3$, $t+6$, and $t+7$). 
On the other hand, our proposed solution, which combines the CNN and ViT architectures for feature extraction and a GRU time-series model for prediction, is able to learn more complex relationships between the inputs and outputs, thanks to the important information extracted from the features extraction component about the surrounding environment motion other than only the detected bounding boxes, resulting in a more accurate and robust model.

\subsubsection{Qualitative Evaluation}
\begin{figure}[!t]
    \centerline{\includegraphics[width=1 \columnwidth]{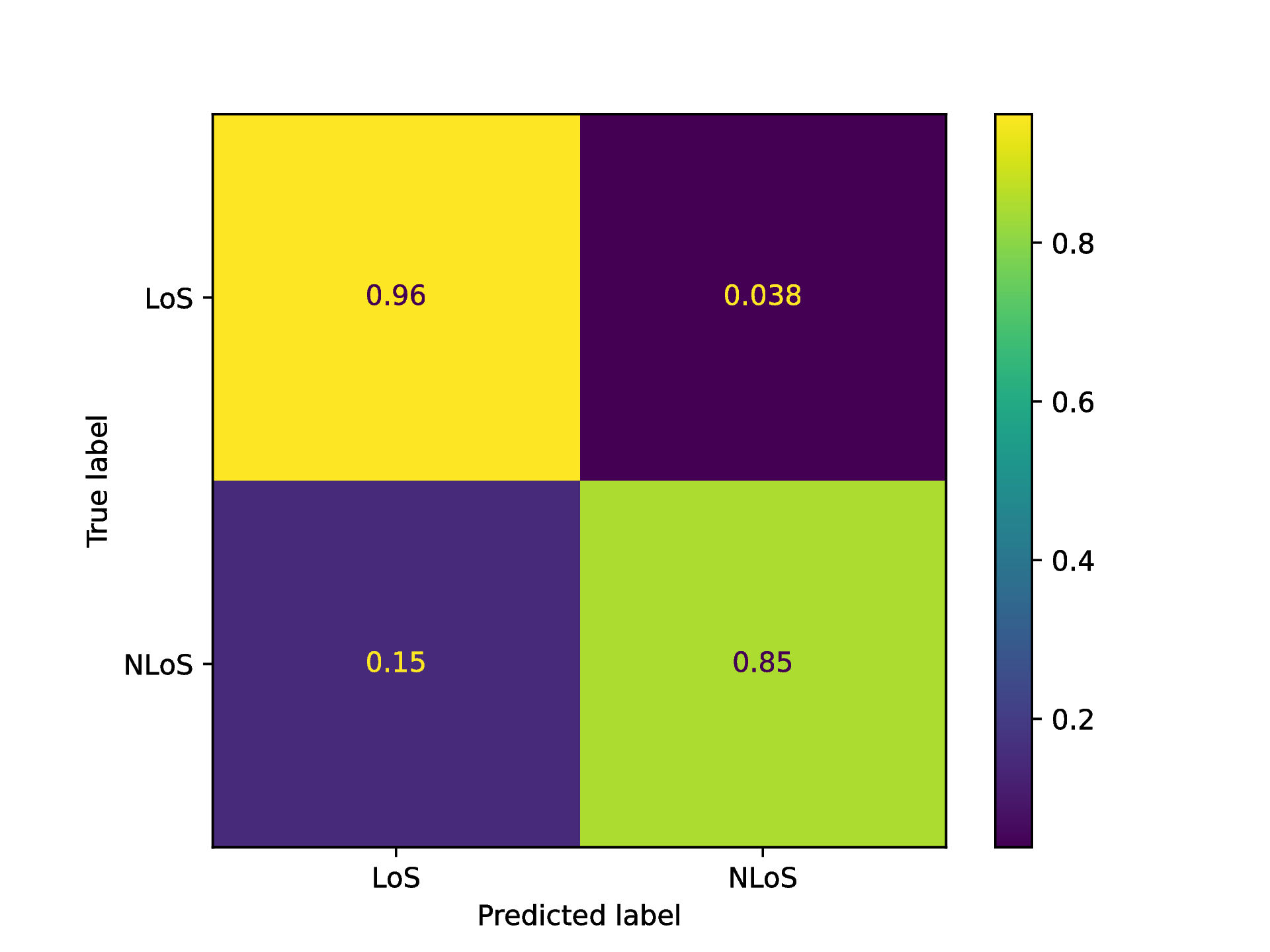}}
    \caption{Confusion matrix of the proposed solution.}
    \label{corr}
\end{figure}

To assess the precision of the model, we depict the confusion matrix, which provides a detailed breakdown of the model's classification performance. 
As observed in Figure \ref{corr}, the confusion matrix shows that the model achieves a True Positive (TP) value of $85\%$ and a True Negative (TN) value of $96\%$.

We also calculate the precision and recall of the proposed method. 
The precision measures the proportion of predicted positives that are correctly positive, while the recall measures the proportion of actual positives that are correctly predicted as positive.
The results show that the proposed methodology achieves a precision of $95.72\%$ and a recall of $85\%$, which proves the effectiveness and robustness of the proposed approach. However, there is an imbalance in the prediction performance, which is resulting from the imbalance of our dataset which has more LoS data points than NLoS ones. This limitation will be considered in our future works.

\begin{figure}[!t]
\centering
\subfloat[SNR]{%
  \includegraphics[width=1\linewidth]{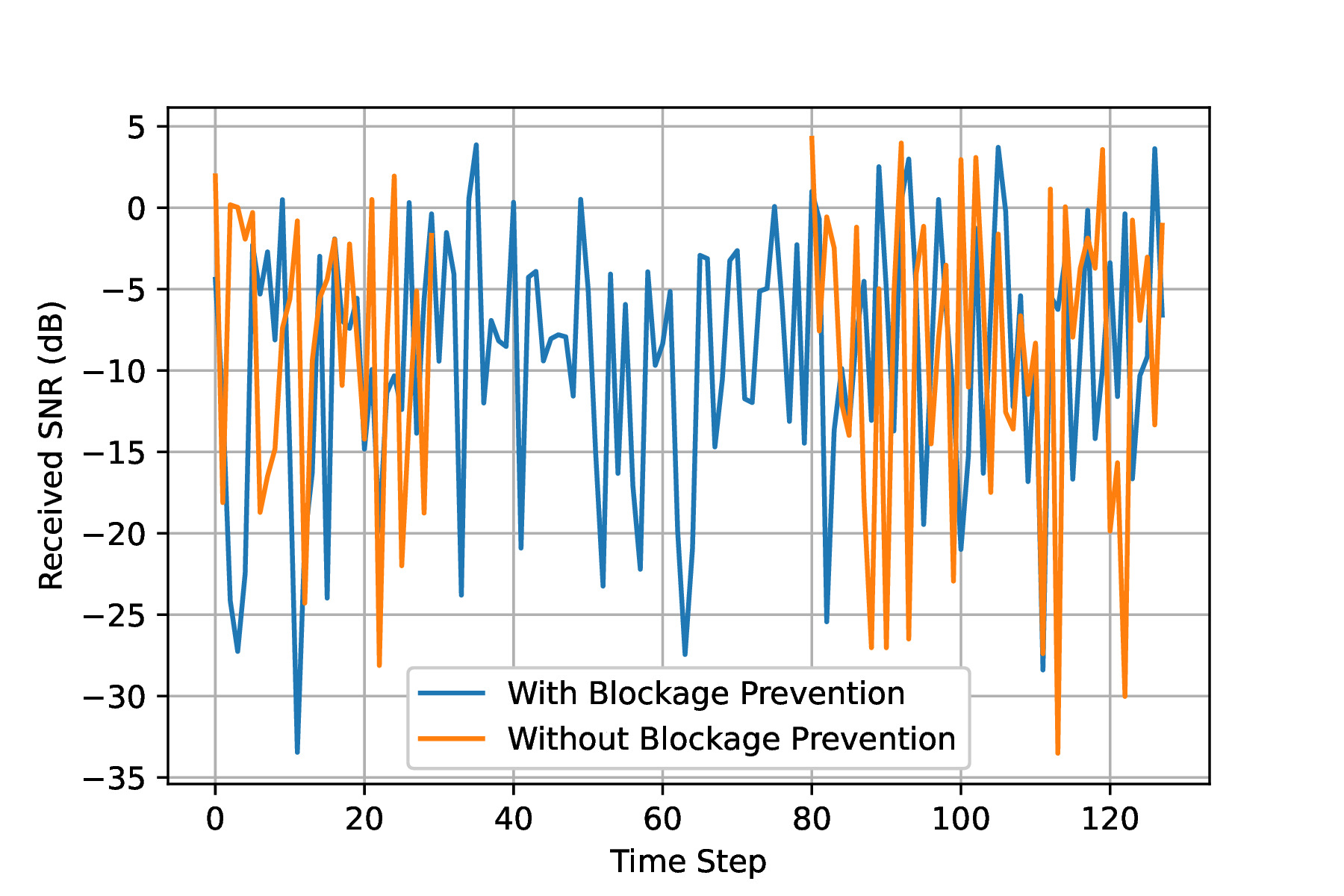}%
}
\hfil
\subfloat[Channel Capacity]{%
  \includegraphics[width=1\linewidth]{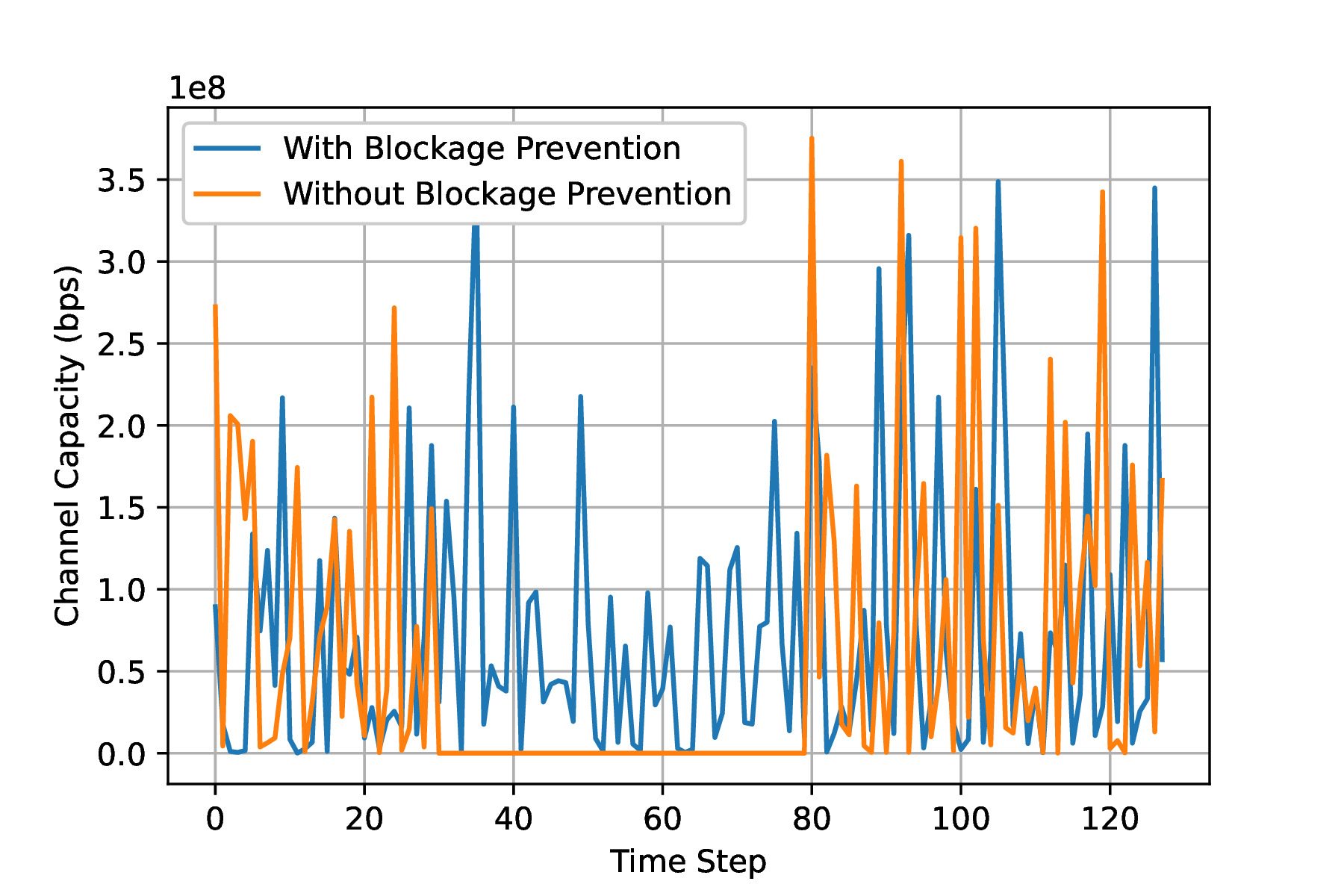}%
}
\caption{Evolution of (a) SNR and (b) channel capacity for one user with and without blockage prevention.}
\label{snrc}
\end{figure}

\begin{figure}[!t]
\centering
\subfloat[$p$]{%
  \includegraphics[width=1\linewidth]{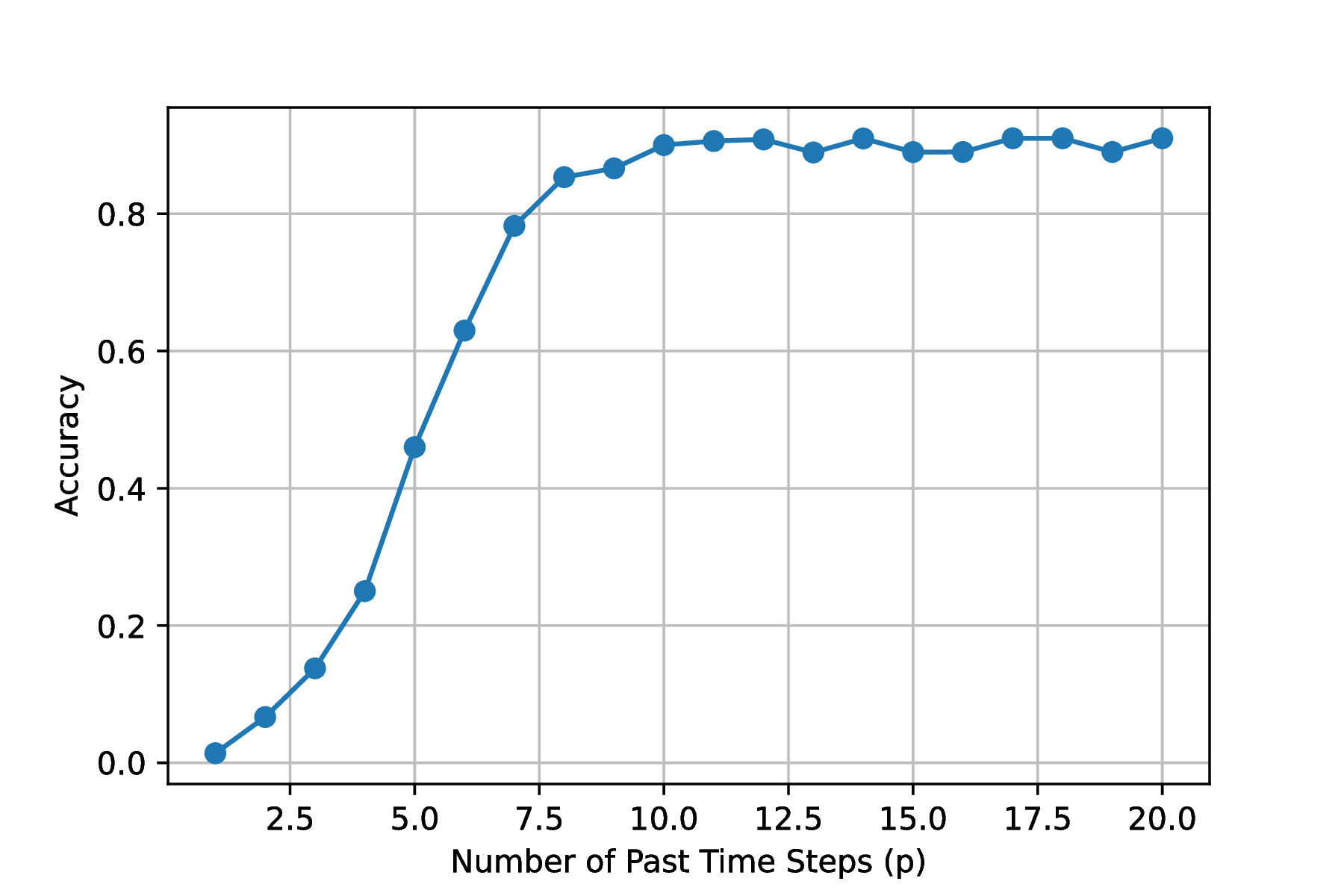}%
}
\hfil
\subfloat[$f$]{%
  \includegraphics[width=1\linewidth]{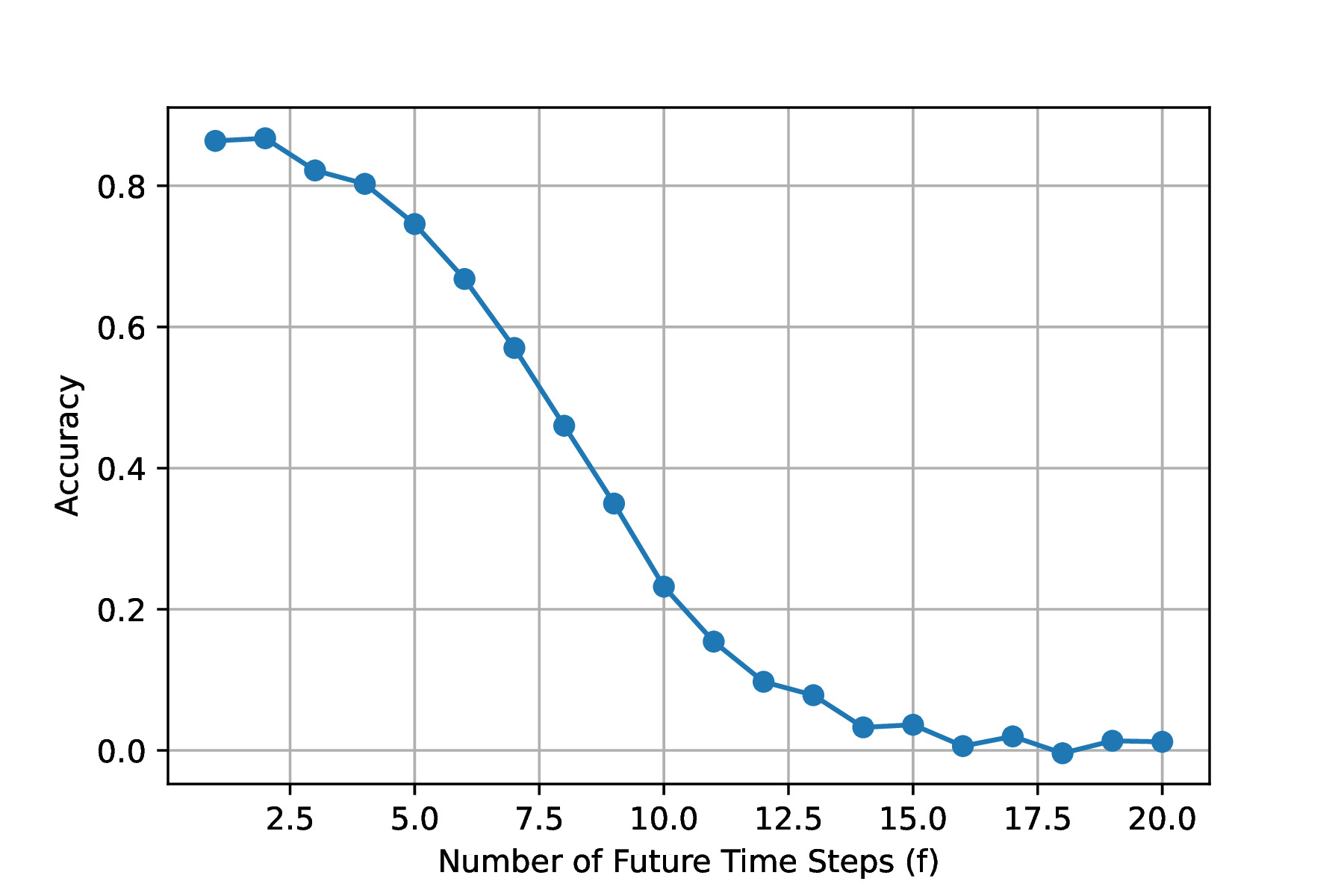}%
}
\caption{Evolution of accuracy with changing number of (a) past time steps $p$ and (b) future time steps $f$.}
\label{pf}
\end{figure}

To further evaluate the performance of our proposed solution for blockage detection, we conducted two distinct scenarios. In the first scenario, we simulate a normal V2X communication without any blockage prevention technique and evaluate the performance in terms of SNR, channel capacity, and accuracy (as shown in Figure \ref{snrc}).
From the result illustrated in Figure \ref{snrc} (a), the SNR value exhibited a significant drop when a blockage occurred between 30 and 80 time steps, leading to a temporary disconnection. Furthermore, the channel capacity plot showed a complete disconnection during blockage, with the communication receiving 0 bps in Figure \ref{snrc} (b). This demonstrates the disruptive impact of blockages on V2X networks, resulting in user disconnections and degraded service quality.
In contrast, in the second scenario, we implemented a blockage prevention technique using our blockage detection framework. When a blockage was detected, the communication was efficiently handed over from base station 1 to base station 2, effectively preventing the SNR drop and maintaining continuous communication. The SNR plot in Figure \ref{snrc} (a) shows that our solution successfully prevents the SNR cut, allowing the communication to persist normally. Similarly, the channel capacity plot in Figure \ref{snrc} (b) shows that the channel capacity remained stable and continued to evolve during the time of blockage, ensuring uninterrupted data flow. This compelling evidence highlights the importance of accurately predicting blockages in advance and the significant role our solution plays in preventing user disconnections and ensuring the stability and reliability of V2X network management. Our proposed architecture's ability to predict future blockages and implement prompt preventative actions presents a crucial advancement in V2X communication, enhancing safety, efficiency, and overall performance.

The selection of the number of past time steps $p$ and future time steps $f$ in our GRU time series model is a critical consideration for accurate predictions. To understand their impact, we conducted experiments varying $p$ and $f$ from 1 to $20$ as shown in Figure \ref{pf}. The results revealed interesting insights into the relationship between these parameters and the model's accuracy.
For $p$, increasing its value generally benefits time series models, particularly those based on recurrent neural networks like GRU. By increasing $p$, the model gains access to more historical context, enabling it to capture long-term dependencies and patterns in the data more effectively. Our experiments showed a notable improvement in accuracy as we increased $p$ from 1 to 8, indicating that an 8-time step history is optimal for capturing the essential patterns. However, beyond this point, the model's performance started to plateau, suggesting that excessively long histories may introduce noise or irrelevant information, hindering further accuracy gains and even leading to performance degradation, as well as increasing computational complexity.
For $f$, we observed that increasing $f$ beyond a certain point led to a decline in accuracy. For our model, a value of 3 for $f$ appeared to strike the right balance, providing a reasonable prediction horizon without overburdening the model with excessive future information.

These findings indicate the importance of carefully selecting the values of $p$ and $f$ to optimize the model's accuracy and avoid unnecessary computational overhead. Our experiments provide valuable insights into configuring the GRU time series model effectively, enhancing its predictive capabilities for future blockage of LoS.

Overall, the numerical results demonstrate the effectiveness and robustness of the proposed methodology in predicting future obstacles in vehicular communications and highlight the potential for further improvements and applications in real-world settings.

\section{Conclusion}
In this paper, we present a novel ViT-based feature extraction architecture to solve the challenging problem of predicting blockages in an mmWave vehicular network.
The proposed architecture enables the extraction of relevant information from the input multimodal data, leading to more accurate and robust predictions about future LoS states using a GRU time-series model as a predictive component.

In contrast to previous approaches that rely on other computer vision techniques, such as object detection and bounding boxes, our method takes a unique perspective by interpreting an image as a sequence of patches and processing it using a standard transformer encoder. This novel approach has demonstrated remarkable performance,  achieving comparable or even superior results to state-of-the-art methods.

The results demonstrated that our proposed approach outperforms the state-of-the-art method \cite{9512383} by achieving a high accuracy rate and proving the effectiveness of the proposed feature extraction component in providing the needed information to the DL time-series model.
Our findings offer valuable insights into the design and implementation of more efficient and reliable wireless networks, which will provide uRLLC and eMBB services in real-world vehicular communication networks and enable the realization of the full potential of 6G V2X applications.

As a future work, it would be interesting to explore more complex vehicular environments and different scenarios, such as urban scenarios, where the presence of obstacles and moving vehicles can significantly affect the performance of the network. Moreover, it would be valuable to investigate the feasibility of using our proposed feature extraction architecture to extract information from other input data types, such as Lidar data, and explore their potential to enhance the prediction accuracy of the network.

\bibliographystyle{IEEEtran}
\bibliography{IEEEabrv,Bibliography.bib}

\end{document}